\documentclass[lettersize,journal]{IEEEtran}
\usepackage{amsmath,amsfonts}
\usepackage{algorithmic}
\usepackage{algorithm}
\usepackage{array}
\usepackage[caption=false,font=normalsize,labelfont=sf,textfont=sf]{subfig}
\usepackage{textcomp}
\usepackage{stfloats}
\usepackage{url}
\usepackage{verbatim}
\usepackage{graphicx}
\usepackage{cite}
\usepackage{multirow}
\usepackage{amsmath,amssymb,amsfonts}
\usepackage{textcomp}
\usepackage[dvipsnames]{xcolor}
\usepackage{booktabs}
\usepackage{mdwtab}
\usepackage{multirow}
\usepackage{caption}
\usepackage{soul}
\sethlcolor{yellow}
\newcommand{\bm}{\textcolor{black}}
\newcommand{\redm}{\textcolor{black}}
\newcommand{\grnm}{\textcolor{black}}
\newcommand{\prpm}{\textcolor{black}}
\newcommand{\orm}{\textcolor{black}}

\usepackage{sistyle}
\SIthousandsep{,}

\begin{document}

\title{Exploring Deep Neural Networks on Edge TPU}

\author{
\IEEEauthorblockN{
Seyedehfaezeh Hosseininoorbin\textsuperscript{*1,2},
Siamak Layeghy\textsuperscript{2},
Brano Kusy\textsuperscript{1},
Raja Jurdak\textsuperscript{3}, 
Marius Portmann\textsuperscript{2}}

\IEEEauthorblockA{\textsuperscript{1}DATA61, Commonwealth Scientific and Industrial Research Organisation (CSIRO), Australia}
\IEEEauthorblockA{\textsuperscript{2}School of Information Technology and Electrical Engineering, The University of Queensland, Australia \\
\textsuperscript{3} Queensland University of Technology, Australia,
\\
{f.noorbin@uq.net.au, siamak.layeghy@uq.net.au, brano.kusy@data61.csiro.au, \\ r.jurdak@qut.edu.au,
marius@itee.uq.edu.au}
}
\thanks{*Corresponding author
\newline{{\it E-mail address:} f.noorbin@uq.net.au (Faezeh Noorbin).}}}




\maketitle

\begin{abstract}
\bm{This paper explores the performance of Google's Edge TPU on feed forward neural networks. We consider Edge TPU as a hardware platform and explore different architectures of deep neural network classifiers,} which traditionally has been a challenge to run on resource constrained edge devices. 
\bm{Based on the use of a joint-time-frequency data representation, also known as  spectrogram, we explore the trade-off between classification performance and the energy consumed for inference. 
The energy efficiency of Edge TPU is compared with that of widely-used embedded CPU ARM Cortex-A53. 
Our results quantify the impact of neural network architectural specifications on the Edge TPU's performance, guiding decisions on the TPU's optimal operating point,} where it can provide high classification accuracy with minimal energy consumption. \bm{Also, our evaluations highlight the crossover in performance between the Edge TPU and Cortex-A53, depending on the neural network specifications.  Based on our analysis, we provide a decision chart to guide decisions on platform selection based on the model parameters and context.}
\end{abstract}


\section{Introduction}\label{intro}
 
\bm{Artificial Intelligence (AI) and more specifically machine learning (ML) have gained significant attention from researchers and industry in recent years for solving complex prediction and classification problems. A sizeable portion of the data that is used as input to ML models originates from the Internet of Things (IoT), which has also seen sizeable growth in the number and types of devices deployed globally. The confluence of the two technologies will enable the Intelligent Edge and will result in a significant transformation of pervasive systems  towards more decentralised and in-situ analytics.} 
While in most cases ML models will be trained in the cloud, it will be increasingly necessary to run the inference tasks at the edge, due to bandwidth, latency, cost or privacy constraints that prevent the raw sensor data from being transmitted to the cloud. 
Running advanced ML algorithms on resource-constrained edge devices has been a major challenge. This is particularly so for deep neural network (DNN) models due to their demanding computational, memory and energy footprint. 
As a result, Google, NVIDIA and others have recently developed dedicated hardware accelerator platforms to support ML inference at the edge. However, the performance-energy trade-off of deep learning on these platforms is currently not well understood, which is what we investigate \bm{in our recent preliminary work~\cite{myPerCom2021} and is extended in this paper.}

We specifically explore Google's Edge TPU~\cite{edgetpu}, which has the ability to perform 4 trillion operations per second (TOPS) consuming only 2~Watts of power.
In particular, we explore the Edge TPU's ability for deep learning at the edge, 
\bm{by varying different feed forward neural network hyper parameters including, the number of input nodes, number of hidden layers and nodes at hidden layers.
In order to be able to explore the impact of  the number of input nodes without changing the number of input features, we use the joint-time-frequency data representation, in particular spectrogram, \bm{in this exploration}.}

\bm{We consider cattle activity classification, using a realistic large scale dataset, as a motivating use case.  Within this context, we convert the activity time series to spectrogram representation.}
%
%
%
We demonstrate that this representation has excellent scaling properties, and can significantly compress data, and hence the neural network model, while maintaining a high predictive power and classification performance.

As a key contribution of the paper, \bm{we perform a systematic analysis of the Edge TPU performance and model architectural trade-off by studying} different scales of the model via different spectrogram resolutions
\bm{and hyper-parameters of the neural network architecture, i.e. scaling neural networks with different model sizes and number of parameters, }
on different ML hardware and software platforms, and explore the impact on ML performance (classification $F_1$ Score) and energy efficiency. 
\bm{The practicality of Edge TPU is further studied by implementing the same experiments on two compute platforms, ARM Cortex-A53 which is a power-efficient 64-bit embedded CPU and a traditional Intel i7-4790 CPU.}

Our experimental evaluations reveal a number of interesting findings. 
The results show that the Edge TPU can perform exceptionally well compared to traditional CPUs, with comparable classification performance while significantly reducing energy cost, which is critical for edge ML applications. 
\grnm{We  also find that achieving the optimal performance from the Edge TPU, both in terms of inference performance and 
energy efficiency is very sensitive to the choice of model size  and number of parameters. The energy efficiency of the Edge TPU can dramatically degrade, if the model size or number of parameters are not within the optimal operating region.}

\bm{Surprisingly, Cortex-A53 is more energy efficient than Edge TPU for very small model sizes.
Our evaluations show that there is a relatively narrow ``sweet spot" in terms of model size  and number of  parameters, within which Edge TPU achieves high classification performance and very high energy efficiency. All configurations outside of this sweet spot lead the Cortex-A53 to again outperform the  Edge TPU.  
To the best of the authors' knowledge, this paper is the first to report }this highly non-linear behaviour of the Edge TPU.

\orm{Based on our extensive analysis, we provide a decision chart and associated guidelines  for selecting the most suitable platform and for tuning the ML model size and structure} to achieve the optimal performance from the selected platform, both in terms of inference performance and energy efficiency.

The rest of the paper is organised as follows. 
Key related work is briefly summarised in Section~\ref{lit}. 
Section~\ref{use-case} presents the use case application of cattle activity classification, the dataset, and we briefly discuss the joint-time-frequency domain representation (spectrogram).
%
%
\bm{In Section~\ref{methods} our experimental methodology is presented, including spectrogram scaling, and DNN scaling, and we briefly provide background on Edge TPU and Cortex-A53, and Intel i7 platforms. 
%
%
%
%
Results are presented in Section~\ref{Results}, including the classification performance Section~\ref{res-F1}, the energy efficiency Section~\ref{res-energy}, the ratio of Cortex-A53 and Edge TPU Section~\ref{res-ratio}. 
In Section~\ref{guideline}, a decision chart and guidelines on model design for the Edge TPU are provided; and is}
followed by conclusions in Section~\ref{conclusion}.

\section{Related Work}\label{lit}
Machine learning at the edge has recently attracted significant attention, both in industry-based and academic research~\cite{Kumar2020,Merenda2020}. 
A number of ML hardware accelerator platforms, such as Google's Edge TPU, have recently been developed, and a number of papers have evaluated them.
For example, Wisultschew et al.~\cite{Wisultschew2019} studied and compared the performance and efficiency of the Google's Edge TPU and the Intel's Movidius Neural Compute Stick (NCS) for 3D-object detection. 
Hui et al.~\cite{RN830} compared the accuracy in 3D object detection of different ML hardware acceleration platforms, i.e. Google Edge  TPU, NVIDIA Xavier, and NovuTensor. 
Reuther et al.~\cite{eTPUvsInteli9} compared the performance of the Edge TPU to an Intel\textsuperscript{\tiny{\textregistered}} Core\textsuperscript{\tiny{TM}} i9-9900k CPU, as well as and Intel's Neural Compute Stick 2 (NCS2). As a key result, the paper reports a similar inference performance of Edge TPU compared to the i9-9900k CPU, but with, not surprisingly, a significantly lower power consumption. 
\bm{Kljucaric et al.~\cite{Kljucaric} compared the performance and efficiency of NVIDIA Xavier, Edge TPU, and NCS2 for optical character recognition using AlexNet and GoogleNet. The authors reported while NCS2 is more efficient for AlexNet, Edge TPU outperforms with GoogleNet.}

\bm{In~\cite{myNIDS_Arxiv}, the architectural trade-offs between computational and energy efficiency for both feed forward and convolutional neural networks are explored on Edge TPU and Cortex-A53. The authors proposed a deep learning-based network intrusion detection system at the edge for IoT networks using a time-series dataset with fixed number of input features. It is demonstrated that for both types of neural networks, the model size is the determinant parameter for Edge TPU's performance.}

\bm{Later, the performance of the Edge TPU accelerators is evaluated extensively for 423K different convolutional neural networks using NASBench dataset by Google's research team~\cite{edgetpu2021}. The latency and accuracy of different model structures are studied on 3 different configurations of the Edge TPU. However, the Edge TPU configurations are not adjustable on user-level yet. The authors also proposed a graph-based neural network model to predict the performance of the models based on the given structure.}

\bm{Our recent preliminary work~\cite{myPerCom2021} studies the performance of feed-forward neural networks with 4 hidden layers for different number of input nodes on Intel i7 and the Edge TPU. 
The focus was on the exploration of the spectrogram data representation and its scaling properties.
We demonstrated that spectrogram data representation allows the compression of the data while maintaining its predictive power.
}

To the best of our knowledge, none of the related works have explored \bm{the details of the Edge TPU's performance in terms of its sensitivity to the architectural model variations and model specifications, which is the focus of this paper. 
}
In order to scale the neural network model size, we use the scalability properties of the joint-time-frequency domain data representation, and present the results of our extensive experimental evaluations. These evaluations were performed in the context of a cattle activity classification application, which is discussed in Section~\ref{use-case}.

 \section{Use Case: Cattle Activity Classification}\label{use-case}

As a use case, we consider the problem of automated cattle activity classification, which is of increasing interest to the beef and dairy industry~\cite{RN256, RN247}.
The quantity and quality of beef and dairy produce are directly related to the animals' health and welfare, which in turn can be inferred from animal behaviour ~\cite{RN618}.
It has been demonstrated that deep neural network classifiers have a great potential for this application~\cite{myPhdForum,myCompAG2021}.
While training of the neural network classifier can easily be performed in the cloud, the actual activity classification (inference) needs to be done at the edge, due to limited, intermittent or expensive network connectivity (satellite link) of the embedded sensor devices that are attached to the animals.
An example of such an edge device is the Ceres Tag~\cite{RN808}, a small solar powered device attached to the ear of cattle, which is equipped with multiple sensors and satellite communication. 

Due to computational and energy requirements, it has traditionally been a challenge to run deep learning algorithms on very resource constrained edge devices. 
\orm{This paper explores Edge TPU through the motivating application of cattle activity classification.} 
 \bm{In particular, we explore how scaling of the neural network architecture
impacts on the performance of Edge TPU. These explorations are applicable to different fields of study which interest in exploring the DNN hyper-parameters, and using and evaluating Edge TPU ASIC~\cite{myNIDS_Arxiv,myCompAG2021,myPerCom2021}.}
We will use the joint-time-frequency domain sensor data representation (spectrogram) to \orm{ be able to scale the input size of the models},
since its potential in this application context has been demonstrated in our earlier work~\cite{myPhdForum,myCompAG2021,myPerCom2021}
\footnote{We note that, while the power requirements of Edge TPU are still higher than what an ear-tag sized solar power cell can currently support, the use of Edge TPU is feasible on larger, collar-sized sensors.}.



 \subsection{Activity Classification Architecture}\label{scenario}
 
  \begin{figure*}[!t]
 \centering
 \includegraphics[width=\textwidth]{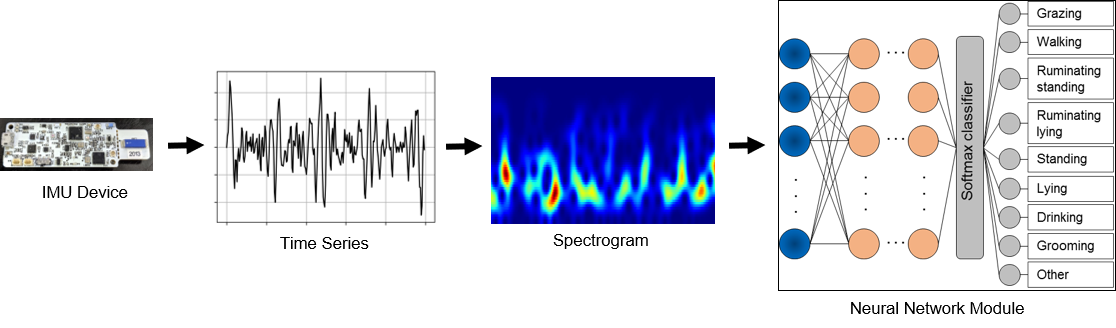}
 \caption{\bm{A schematic illustration of the activity classification architecture.}}
 \label{fig:workflow}
 \end{figure*}

 An overview of our activity classification architecture, which is discussed in more detail in~\cite{myCompAG2021}, is shown in Figure~\ref{fig:workflow}. 
%
%
The classification process starts on the left with the collection of 
time series from a tri-axial IMU (Inertial Measurement Unit) sensor that is attached to the subject. 
%
%
The data is then converted into a joint-time-frequency representation (spectrogram), 
and fed into a feed-forward Deep Neural Network (DNN) 
to return one of the 9 considered activity classes, as shown in the figure. 


 
 \subsection{Background: Joint Time-frequency Representation}\label{Bck-spec}

 \begin{figure}[!b]
 \centering
 \includegraphics[width=9.4 cm]{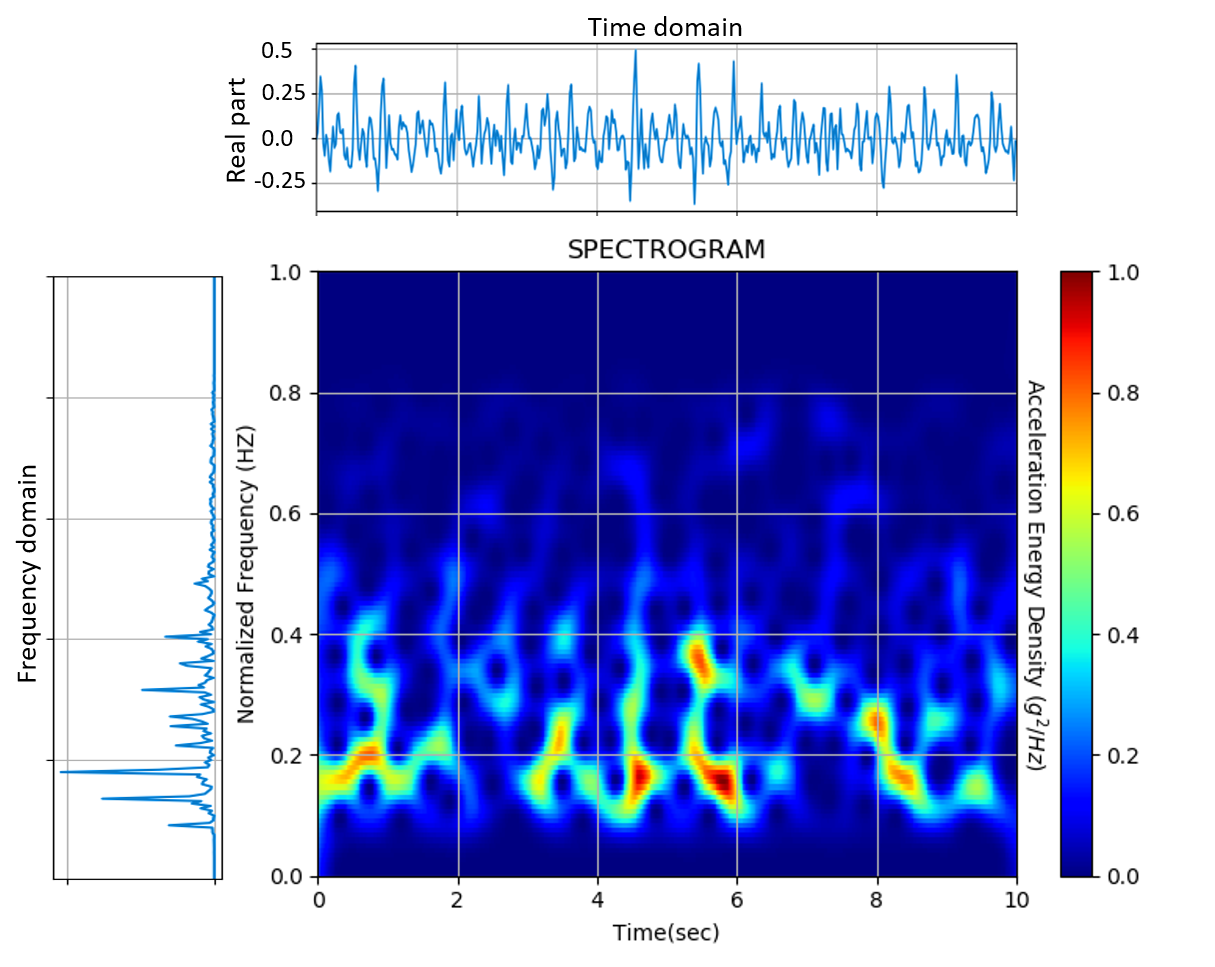}
 \caption{\bm{A sample spectrogram for a 10 second window of cattle activity acceleration signal, with the signal also shown in their respective time-domain and frequency-domain representations.}}
 \label{fig:SpecTF}
 \end{figure}
 \bm{
 Many human and animal related activity monitoring studies have been performed by recording and analysing accelerometery signals. 
 In some of these activity classification studies it has  been shown that the human activity acceleration~\cite{Ravi2017}, human fetal activity acceleration~\cite{layeghy2014a,layeghy2014b}, and cattle activity acceleration signals~\cite{myCompAG2021} have a time-varying spectral content.}
 
 While such signals are traditionally represented in the time or frequency domain, a joint-time-frequency representation of these signals has much greater potential in the context of activity classification~\cite{myCompAG2021,Ravi2017}.
 %
A commonly used method for joint-time-frequency data representation is a \textit{spectrogram}. 
The spectrogram representation of a discrete time signal $x[n]$ is computed as shown in Equation~\ref{formula:d-STFT2}, with $n$ and $\omega$ representing discrete time and frequency respectively, and with $W[n]$ representing the windowing functions~\cite{Ravi2017}.

  \begin{flalign}
 S{_W}[n,\omega]&=|\sum_{m=-\infty}^{\infty} x[n]W[n-m]e^{-j\omega m}|^2
 \label{formula:d-STFT2}
 \end{flalign}

 Figure~\ref{fig:SpecTF} shows an example spectrogram of an acceleration sensor signal, obtained from a sensor placed on cattle for the purpose of activity classification. Our use case application scenario is discussed in more details in Section~\ref{scenario}.
The horizontal and the vertical axes represent time and normalised frequency respectively. In this figure, colour indicates acceleration energy density (normalised), ranging from red (high) to blue (low). For illustration purposes, the corresponding acceleration signals in time-domain and frequency-domain representation are shown at the top and left respectively.

 \subsection{Dataset}\label{dataset}
  The dataset we used for our experiments was collected by CSIRO in Armidale, Australia, over a one month period in 2018. The data was obtained from tri-axial IMU (Inertial Measurement Unit) sensors that that were attached via collars to 10 cattle.
  The sensors collected accelerometer, magnetometer and gyroscope readings in 3 axes, resulting in 9 streams of time-series sensor data sampled at 50 Hz, and with a total of more than 3.5 million samples.
 The dataset was labelled manually (by human recorders) with the corresponding 9 cattle activity classes. Further details on the dataset are available in~\cite{myCompAG2021}.

\section{Experimental Methodology}\label{methods}


\grnm{In this study, to evaluate the Edge TPU performance, we explore the DNN hyper-parameters with the focus on varying the number of nodes at input layer ($n_{x}$), number of hidden layers ($L$), and the number of nodes at hidden layers ($n_{h}^{[l]}$).
Since the model size corresponds to the input layer size (spectrogram resolution) and number of hidden layers and their number of nodes,
changing the considered hyper-parameters leads to scale the model size and number of parameters per layer.  
In the following sections, the spectrogram scaling and DNN architecture scaling are explained.}

\subsection{Spectrogram Scaling}\label{spec_scaling}

 \begin{figure}[!b]
 \centering
 \includegraphics[width=6.0cm]{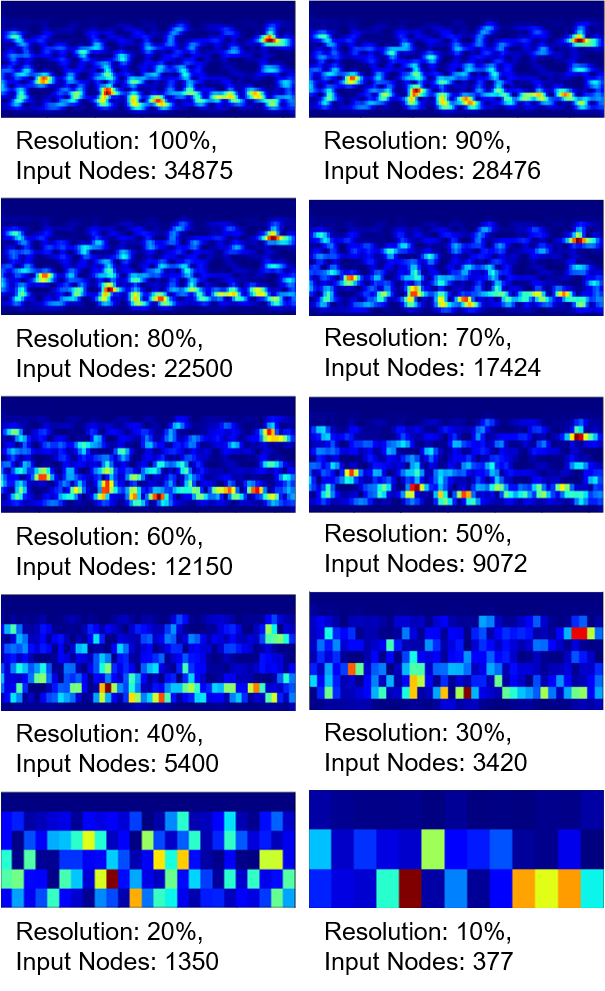}
 \caption{\bm{The different steps of scaled resolution for an example spectrogram.}}
 \label{fig:Resdim}
 \end{figure}

 \orm{In order to match the input size to the varied number of nodes at input layer, we use the scaling properties of 
 the spectrogram data representation.}
 %
 The goal is to explore different scales (or resolutions) of the spectrogram and evaluate the impact on the model size, classification performance and energy consumption. \orm{The focus is on the model architectural variation
 and energy efficiency trade-off, which is crucial for resource-constrained edge devices.}
 
 %
 In this context, we define spectrogram resolution as the ratio of the one-dimensional pixel count compared to the original (full) resolution.
 A resolution of 50\% refers to the case where the number of pixels in each dimension is halved, which results in a spectrogram image with one quarter of the size, i.e. number of pixels. The lowest considered resolution of 10\% has therefore a size of only 1\% of the original resolution. 

 \bm{In our experiments, we consider spectrograms for 9 axis collected from three tri-axial sensors. A spectrogram with 100\% resolution leads to 3875 nodes per axis and in total 34875 nodes at the input layer of the DNN.}
 In our experiments, we consider spectrogram resolutions in steps of 10\%, i.e. from 10\% to 100\%.
 The scaled versions of the spectrograms are computed via bicubic downsampling~\cite{bicubic}.
 %
  %
  As an illustration, Figure~\ref{fig:Resdim} shows the different resolutions of an example spectrogram (10 second window) of the accelerometry signal for the ``Ruminating Lying" activity, which is one of the 9 cattle activity classes that we are considering.
%

\subsection{Neural Network Architecture and Scaling}\label{dnn-scaling}
 
 \begin{figure}[!t]
 \centering
 \includegraphics[width=8 cm]{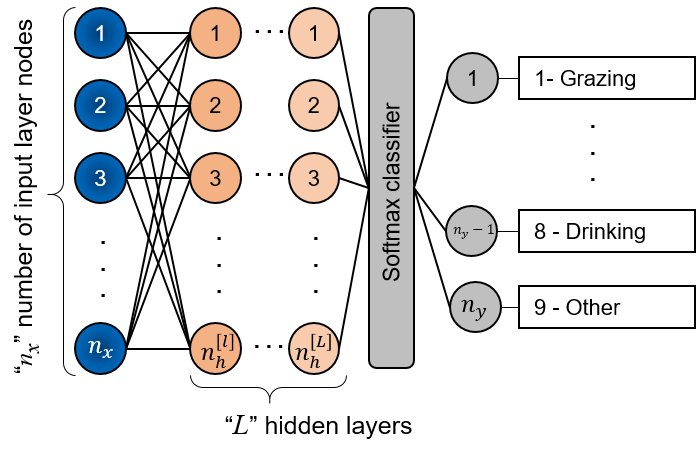}
 \caption{\bm{The neural network architecture.}}
 \label{fig:DNN}
 \end{figure}
 
\grnm{Figure~\ref{fig:DNN} shows a scheme of the feed forward neural network architecture used in this study. 
It is built up by a stack of fully connected layers including the input layer, various number of the hidden layers $L$ and the output layer. 
The number of nodes at the first (input) layer $n_{x}$ equals the size of the input spectrogram, which is determined by its resolution. 
The hidden layers have $n_{h}^{[l]}$ nodes with Rectified Linear Unit (ReLU) activation function for all nodes, except for the last layer $n_{h}^{[L]}$ with 32 nodes. 
The output layer uses the Softmax function, to return one of the 9 considered activity classes, as shown in Figure~\ref{fig:DNN}. 
The loss function for this model is Sparse Categorical Cross-entropy.}

\begin{table}[!b]
\begin{center}
  \caption{\bm{The considered hyper-parameters}}
  \label{tab:params}%
  \small
    \begin{tabular}{| l | l |}
    \hline
    Parameter & Definition  \\
    \hline
    $L$     & number of hidden layers \\
    \hline
    \begin{tabular}[l]{@{}l@{}} $n_{x}$=$n_{h}^{[0]}$ \end{tabular}
     & number of nodes at input layer (input size)  \\
    \hline
    \begin{tabular}[l]{@{}l@{}} $n_{h}^{[l]}$ \\ for $l=1$ to ${L - 1}$ \end{tabular} & number of nodes at the $l$\textsuperscript{th} hidden layer  \\
    \hline
    \begin{tabular}[l]{@{}l@{}} $n_{h}^{[L]}$  \end{tabular} & number of nodes at the last hidden layer  \\
    \hline
    \begin{tabular}[l]{@{}l@{}} $n_{y}=n_{h}^{[L+1]}$  \end{tabular} & number of nodes at output layer (output size) \\
    \hline
    \end{tabular}%
    \end{center}
\end{table}%

%

\orm{The design space of the DNN model is explored by varying the range of number of input nodes ($n_{x}$), number of hidden layers ($L$) and nodes at hidden layers ($n_{h}^{l}$).}
\orm{Table~\ref{tab:params} lists the these hyper parameters and their definitions.} 
%



\grnm{In the first set of experiments, for each number of input nodes, i.e. the spectrogram resolution/scale, number of hidden layers (L) is doubled in each step, starting from 2 up to 64. When increasing the number of hidden layers, the number of nodes at hidden layers remains fixed.
Then, in the next set of experiments, the number of input nodes and hidden layers are fixed at $n_{x} = 5400$ and $L=128$, respectively, and number of nodes at hidden layers ($n_{h}^{l}$) are doubled for each half of the hidden layers at each step of the experiments.
Finally, to investigate the sensitivity of Edge TPU to the choice of parameters at hidden layers, we fix $n_{x} = 377$ and $L = 2$, and change the $n_{h}^{[1]}$ from 5000 to 10,000 nodes with steps of 100 which corresponds to the nodes at the first hidden layer.
}

\subsection{Setup: Software and Hardware Platforms}\label{setup-HW}

 \begin{figure}[!t]
 \centering
 \includegraphics[width=8.5 cm]{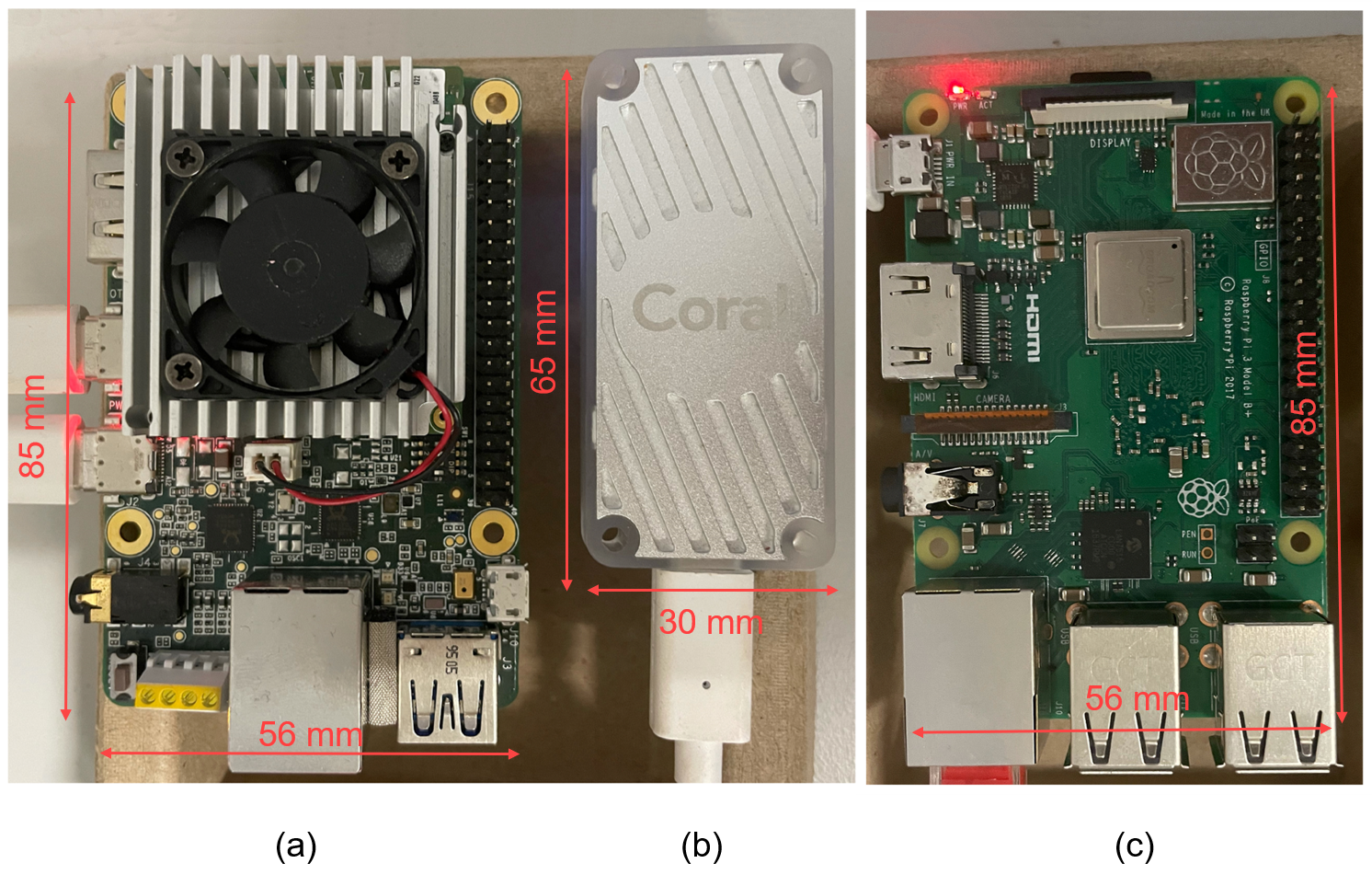}
 \caption{\bm{(a) Coral Dev Board  and (b) Coral USB Accelerator with Edge TPU and (c) Raspberry Pi3 model B+ platforms.}}
 \label{fig:Edge-HW}
 \end{figure}

\bm{While TensorFlow (TF) Lite version 2.5.0 was our main machine learning software platform, different hardware platforms were investigated to study the practicality of Edge TPU. TF Lite is a deep learning framework} aimed at resource constrained embedded and IoT devices, which uses 8-bit unsigned integers instead of 32-bit floating point numbers as in TF~\cite{TensorFlow2015,tflite}.   
\bm{For the hardware platform, while the focus of the paper is the exploration of Edge TPU, other hardware platforms are considered for the comparison which include a common embedded 64-bit CPU and a traditional CPU-based workstation.}
%
%
%

\subsubsection{Google Edge TPU}\label{Bck-TPU}
In 2019, Google launched the Edge TPU, a purpose-built ASIC hardware accelerator for machine learning applications with high performance and a low energy footprint, with the aim of running machine learning inference at the edge~\cite{edgetpu}. 
\bm{Figure~\ref{fig:Edge-HW}-(a) and (b), respectively, show the Coral Dev Board and the USB Accelerator both with Google's Edge TPU.} 
%
The TPU is based on a systolic array architecture, which allows performing matrix multiplication highly efficiently and at a  massive scale. All operations are limited to 8 bit integers, which increases both the performance and energy efficiency~\cite{TPUarch}. 
\bm{In our experiments we used both the Coral USB Accelerator~\cite{usbtpu} and the Coral Dev Board~\cite{devtpu}.}

\subsubsection{ARM Cortex-A53}\label{Bck-A53}

\bm{The first comparison platform is the Raspberry Pi 3B+, which is equipped with a 64-bit quad-core Arm Cortex-A53 CPU @1.4GHz and 1GB RAM. It is powered with a 5V-2.5A power supply running the Raspbian GNU/Linux OS. 
%
Figure~\ref{fig:Edge-HW}-(c), shows the credit card sized Raspberry Pi 3B+.}

\subsubsection{Intel i7}\label{Bck-i7}

\orm{The second comparison platform is a traditional CPU-based workstation using an Intel i7-4790 @3.60~GHz CPU and 64GB of RAM, running the Linux kernel 4.15.}

\section{Results}\label{Results}

\begin{table}[!t]
  \begin{center}
  \caption{\bm{Summary of the considered hyper-parameters to explore the design space of DNN models}}
    \label{tab:set1}%
  \small
    \begin{tabular}{| l | l |}
    \hline
    Parameter & Range \\
    \hline
    $L$     &  \{2, 4, 8, 16, 32, 64\} \\
    \hline
    $n_{x}$ & 
    \begin{tabular}[l]{@{}l@{}}\{377, 1350, 3420, 5400, 9072, 12150,\\ 17424, 22500, 28476, 34875\}\end{tabular}\\
    \hline
    \begin{tabular}[l]{@{}l@{}} $n_{h}^{[l]}$\\for $l=1$ to $L - 1$ \end{tabular}
    &  64 \\
    \hline
    \begin{tabular}[l]{@{}l@{}} $n_{h}^{[L]}$ \end{tabular} &  32  \\
    \hline
    $n_{y}$ & 9  \\
    \hline
    \end{tabular}%
\end{center}
\end{table}%

%
%
\orm{Our results mainly focus on the classification performance, energy efficiency and memory considerations in terms of the architectural variations of the 
neural network models.}
%
\orm{We also consider the ratio of energy efficiency between the Edge TPU and Cortex-A53, which is of critical importance for deploying machine learning on resource constrained edge devices and provides an overall view of the performance of each platform.}


\subsection{Classification Performance vs Model Variations}\label{res-F1}

%
 \grnm{Table\ref{tab:set1} presents the range of  hyper-parameters explored in the first set of experiments. 
 The number of hidden layers is varied from 2 to 64 with the steps of 2 and the number of input nodes are specified in the second row of the Table. 
 %
 The rest of the hyper parameters are fixed as shown in the next rows of the table.}
 
 \begin{figure*}[t!]
 \centering
 \includegraphics[width=\textwidth]{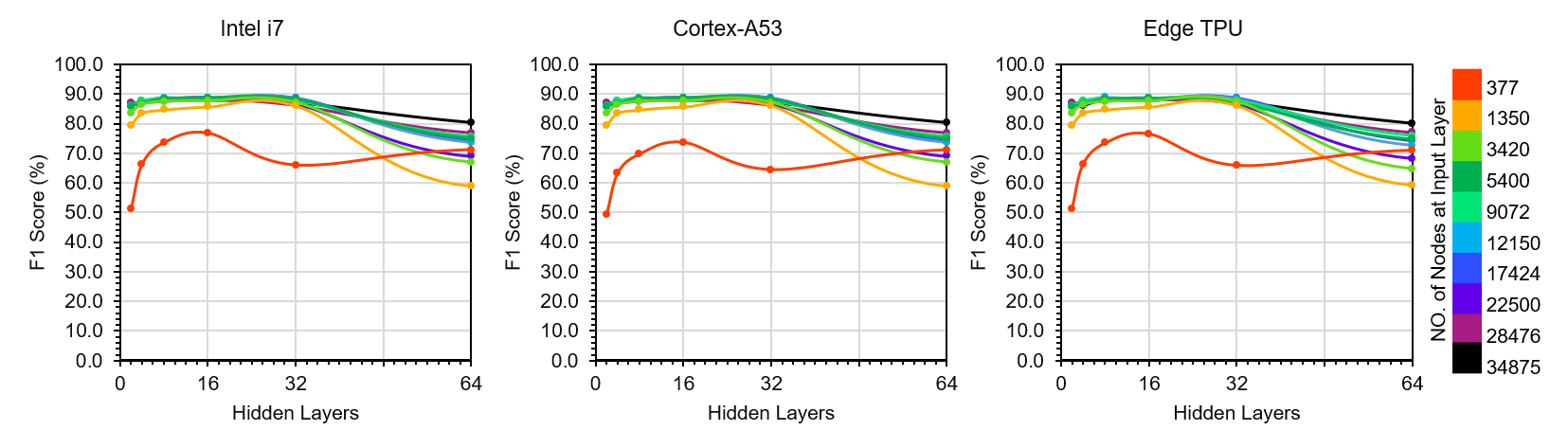}
 \caption{\bm{The average $F_1$ Score versus number of hidden layers for configurations in Table\ref{tab:set1} implemented on i7, Cortex-A53 , and Edge TPU.}}
 \label{fig:F1vsHiddenLayers}
 \end{figure*}

 \grnm{Figure~\ref{fig:F1vsHiddenLayers} shows the $F_1$ Score as a function of number of hidden layers for different number of input nodes. 
 Different colours indicate different number of input nodes as shown in the colour bar. 
 The results are shown for the three HW platforms, and as seen, the F1-Score is very similar between the three platforms.}
 %
\orm{In all cases, except the case with 377 input nodes, the $F_1$ Score first reaches up to 88.8\% with increasing number of hidden layers, and then drops.}
%
 \bm{It can also be seen that the maximum F1 score is achieved with 1350 to 9072 input nodes and 16 to 32 hidden layers.}
 \orm{The $F_1$ Score declines for 64 hidden layers due to the over-fitting to the majority class (our dataset is imbalanced).} 
\begin{figure*}[b!]
\centering
\includegraphics[width=\textwidth]{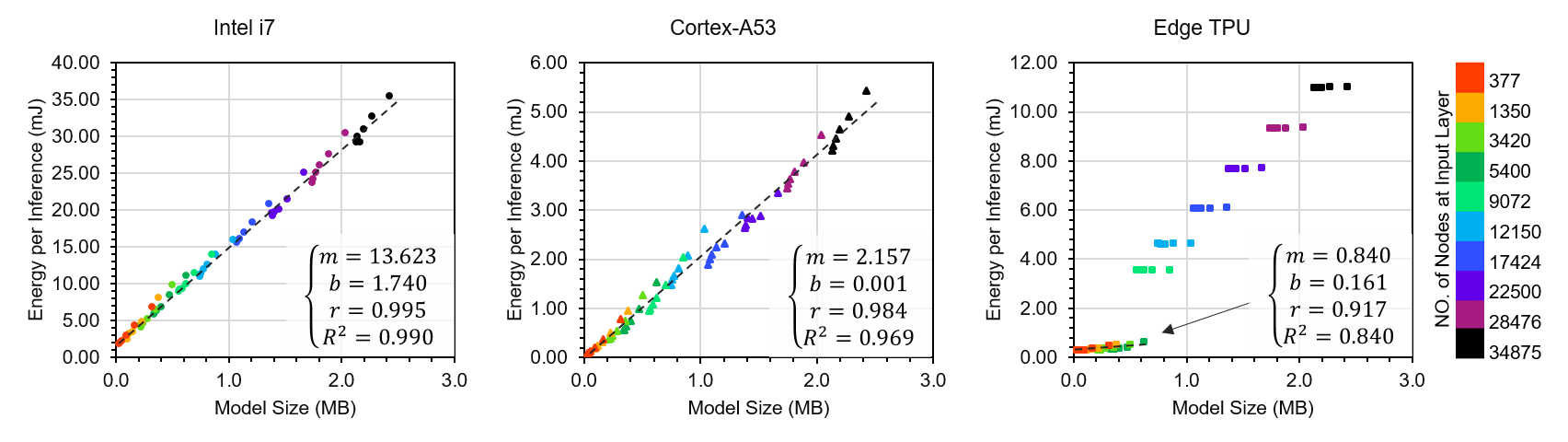}
\caption{\bm{Energy per inference vs model size for configurations in Table\ref{tab:set1} implemented on i7, Cortex-A53 , and Edge TPU.}}
\label{fig:EnergyvsModelSize}
\end{figure*}
 
\subsection{Energy Efficiency vs Model  Variations}\label{res-energy}

\bm{In order to 
estimate the energy per inference for the Edge TPU, we measure the inference time, and multiply it with the peak power consumption of 2 Watts, as specified in~\cite{TPUbench}. 
The same method is considered for the Cortex-A53 CPU on a Raspberry Pi 3B+ with the active bare-board average power consumption of 2 Watts.}
For the measurement of energy per inference for Intel i7, we used the PyRAPL~\cite{pyRAPLdoc} toolkit, which takes advantage of Intel's proprietary ``Running Average Power Limit" (RAPL) technology~\cite{RAPL}.


\subsubsection{Exploring model size}\label{energy-modelsize}

\bm{Figure~\ref{fig:EnergyvsModelSize} shows the inference energy in millijoule (mJ) for a single inference step as a function of model size in MB, which includes the effect of  both the input layer size (spectrogram resolution) and number of hidden layers. 
\orm{The number of input} nodes are identified with different colours. Results are shown \grnm{for the set of hyper-parameters in Table~\ref{tab:set1}} on three HW platforms.} 

\bm{For Intel i7, Figure~\ref{fig:EnergyvsModelSize} shows a relatively linear increase.}
%
\grnm{The slope ($m$) and intercept ($b$) of the regression line equal to 13.623 (mJ/MB) and 1.74 (mJ), respectively. The reliability of the linear proportion is demonstrated by correlation coefficient $r = 0.995$ and coefficient of determination $R^{2} = 0.99$ values.}

In the case of Cortex-A53, Figure~\ref{fig:EnergyvsModelSize}
 shows a qualitatively similar pattern 
\grnm{ where the coefficients of the regression line are as $m = 2.157$ (mJ/MB), $b = 0.001$ (mJ), and $r = 0.984$ and $R^{2} = 0.969$. The slope of the line
is significantly lower compared to Intel i7. }
 
\bm{Finally, Figure~\ref{fig:EnergyvsModelSize} shows the inference energy for Edge TPU platform. Here, we observe a surprising bimodal behaviour, with an almost constant and very low energy usage of 0.29~mJ to 1.05~mJ per inference for the input nodes of 377 to 5400 and the model sizes of 0.03~MB to 0.93~MB. }

\grnm{It consists of one linear (corresponding to input nodes of 377 to 5400) and multiple almost constant sections (corresponding to input nodes of 9072 to 34875). The regression line for the linear section has the following coefficients $m = 0.84$ (mJ/MB), $b = 0.161$ (MB), $r = 0.917$ and $R^{2} = 0.84$.}
\orm{While for the first section (input nodes of 377 to 5400), the energy usage linearly increases with the model size, for the higher number of the input nodes (input nodes of 9072 to 34875) it jumps with the increase of the input nodes, but then stays almost constant with other model size increases (due to the increase of the number of hidden layers).}

%
%

\begin{figure}[t!]
\centering
\includegraphics[width=7.5 cm]{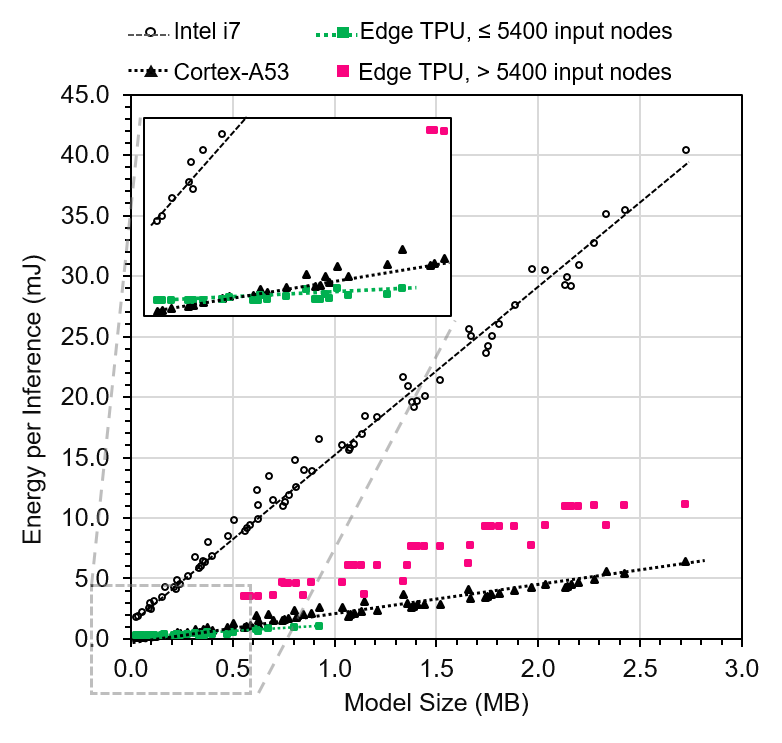}
\caption{\bm{The comparison of energy per inference vs model size on i7, Cortex-A53 , and Edge TPU; focusing on number of input nodes on Edge TPU performance.}}
\label{fig:EnergyvsModelsize_group}
\end{figure}

\bm{Figure~\ref{fig:EnergyvsModelsize_group} shows the three energy consumption graphs from Figure~\ref{fig:EnergyvsModelSize} side by side and  compares the range of energy consumption and behaviour of the three HW platforms. 
With regards to the bimodal behaviour of Edge TPU, two different colours are used to indicate its energy consumption for the 
number of input nodes fewer/greater than 5400.
The zoomed region (lower left) shows that for models with size less than 0.15~MB, Cortex-A53 is more efficient than Edge TPU. 
This cross-over is reported in another paper of ours~\cite{myNIDS_Arxiv} for small models with different neural network configurations on \orm{another application.}}
\bm{Beyond 0.15~MB for input nodes $\leq5400$, Edge TPU is more efficient than Cortex-A53; however, for input nodes $>5400$ Cortex-A53 overtakes Edge TPU again.
}

\begin{table}[b!]
  \centering
  \caption{\bm{Summary of the considered hyper-parameters to investigate the sensitivity of Edge TPU to the model size}}
  \small
    \begin{tabular}{|l|l|}
    \hline
    Parameter & Range \\
    \hline
    $L$     &  128 \\
    \hline
    $n_{x}$ & {5400} \\
    \hline
    \begin{tabular}[l]{@{}l@{}}
    $(n_{h}^{[l_{1}]},n_{h}^{[l_{2}]})$\\
    for ${l_1} = 1$ to $63$\\
    and ${l_2} = 64$ to $127$
    \end{tabular} & 
    \begin{tabular}[l]{@{}l@{}}
    \{(64, 64), (128, 64), (128, 128), \\
    (256, 64), (300, 64), (305, 64), \\ 
    (310, 64), (256, 128), (256, 256), \\ 
    (512, 64), (512, 128)\}
    \end{tabular}\\
    \hline
    \begin{tabular}[l]{@{}l@{}}$n_{h}^{[128]}$
    \end{tabular}&  32 \\
    \hline
    $n_{y}$ & 9 \\
    \hline
    \end{tabular}%
  \label{tab:set2}%
\end{table}%

\orm{To further investigate the bimodal behaviour of Edge TPU, we considered increasing the model size for the largest model before the transition of Edge TPU, with $n_{x} = 5400$ and $L = 128$; and studied the energy consumption and memory usage on the Edge TPU and Cortex-A53.} 
\orm{In this experiment, the model size is increased by increasing the number of nodes at hidden layers. 
The models in this set of experiments have fixed number of input nodes while nodes at hidden layers are doubled in each step, i.e. number of nodes for half of the hidden layers in DNN are doubled on each step.
Table~\ref{tab:set2} summarises the hyper-parameters used in this set of experiments. In this table, $(n_{h}^{[l_{1}]},n_{h}^{[l_{2}]})$ pair shows the number of nodes at hidden layers for layers ${l_1} = 1$ to $63$ and ${l_2} = 64$ to $127$, respectively. The set of pairs shown for the range of  $(n_{h}^{[l_{1}]},n_{h}^{[l_{2}]})$ are used in each step of the experiment.}

\begin{figure}[t!]
\centering
\includegraphics[width=7.5 cm]{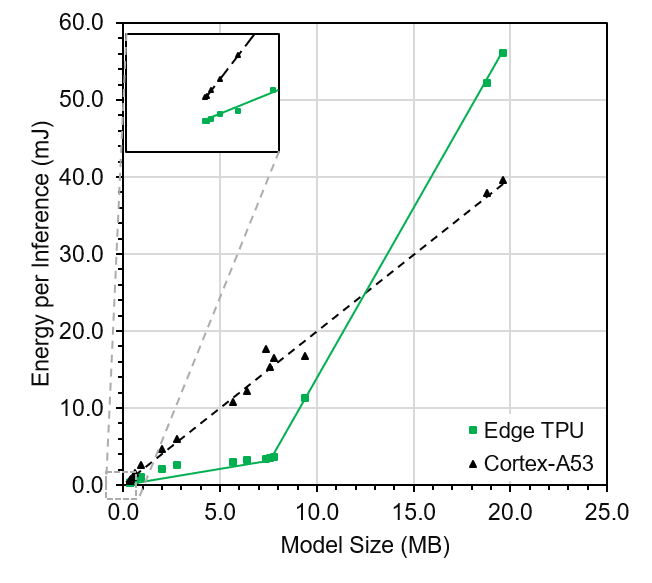}
\caption{\bm{Energy per inference vs model size for configurations in Table\ref{tab:set2} on Edge TPU andCortex-A53.}}
\label{fig:EnergyvsModelsize_R40}
\end{figure}

\orm{Figure~\ref{fig:EnergyvsModelsize_R40} shows the energy usage versus the model size for this experiments. It is shown that up to $\sim8$~MB the energy efficiency is relatively proportional to the increase of the model size. 
The implementation on the Cortex-A53 shows that, the Edge TPU is more efficient for the models with $n_{x} = 5400$ input nodes. However, the advancement of the Edge TPU lasts until the model size is less than $\sim8$~MB before the transition happens. Beyond 8~MB, which is the size of the available on-chip memory, the energy usage curve sharply increases.}

\begin{figure}[b!]
\centering
\includegraphics[width=7 cm]{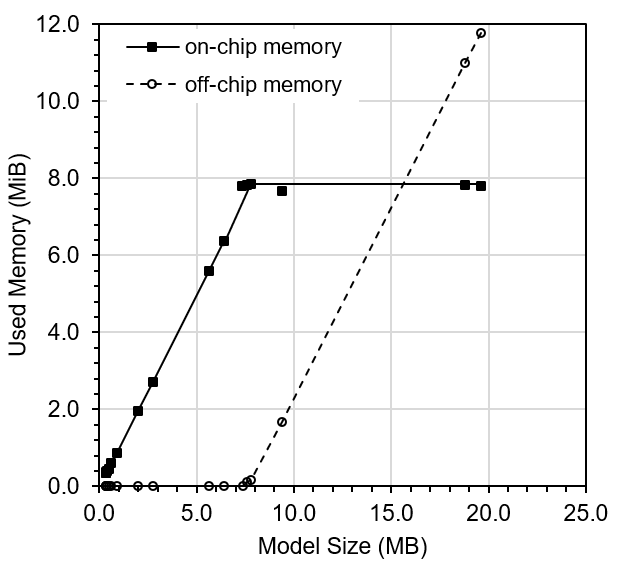}
\caption{\bm{On-chip and Off-chip memory usage vs model size for configurations in Table\ref{tab:set2} on Edge TPU and Cortex-A53.}}
\label{fig:MemoryvsModelsize_R40}
\end{figure}

In order to understand the reason for the transition and cause of the bimodal behaviour, we investigated the compilation reports provided by the Edge TPU compiler~\cite{tpucompiler}, with a particular focus on the memory usage.
\footnote{\orm{The Edge TPU compiler, aims to extract the highest parallelism levels for the execution of the operations when maps the supported operations on the Edge TPU. All the neural networks operations used in this study are supported by the Edge TPU and mapped to execute on chip.}~\cite{edgetpu2021}.}
The Edge TPU uses two main parts of memory for the storage of its model parameters, i.e. on-chip and off-chip memory. 
\bm{
%
Figure~\ref{fig:MemoryvsModelsize_R40} shows the memory usage vs the model size, for the on-chip and off-chip memories in KB. 
As seen, for the models of size beyond 8~MB, the Edge TPU starts using off-chip memory resource to provide the memory requirements of the processes.
This is consistent with the behaviour shown in Figure~\ref{fig:EnergyvsModelsize_R40}, 
where the transition happens for the models of size beyond 8~MB, \orm{which is the amount of available SRAM on Edge TPU.}
}

\subsubsection{Exploring number of nodes per layer}\label{energy-nodes}

\begin{figure}[t!]
\centering
\includegraphics[width=8.9 cm]{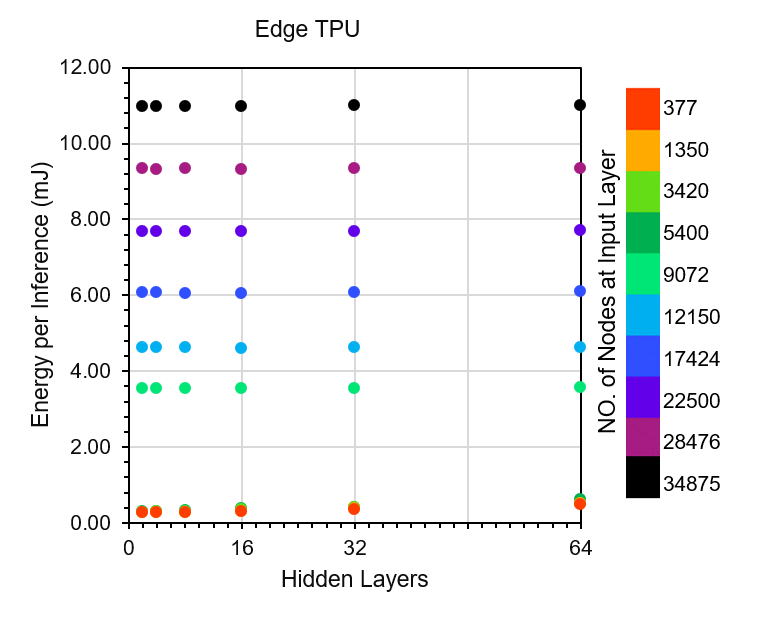}
\caption{\bm{Energy per inference vs number of hidden layers for configurations in Table\ref{tab:set1} on Edge TPU.}}
\label{fig:EnergyvsHiddenLayers_tpu}
\end{figure}

%

\orm{In the last experiment, it is demonstrated that model size is a determinant factor in energy usage, but here we show it is not the only one}. 
\grnm{Taking the note from Figure~\ref{fig:EnergyvsModelsize_group}, demonstrating that model size is not the only determinant parameter of the bimodal behaviour of the Edge TPU regarding the overlap of the energy consumption curves from models with different 
number of input nodes and with different number of hidden layers.}
\orm{As mentioned earlier, model size is affected by both the number of hidden layers and the number of nodes per layer.}
\bm{Figure~\ref{fig:EnergyvsHiddenLayers_tpu} shows the energy consumption of Edge TPU versus number of hidden layers and for different number of input nodes. \orm{It is observed that energy usage is not sensitive to increasing the number of hidden layers. But it is not the case for increasing number of input nodes.}  
For input nodes 377 to 5400, the range and slope of energy usage is almost the same. While for input nodes beyond 5400, 
the energy consumption is incremented in a quantised manner.} 
\prpm{The quantised behaviour of the Edge TPU returns to the HW architecture of this ASIC. }

%
\begin{figure*}[b!]
\centering
\includegraphics[width=16.5 cm]{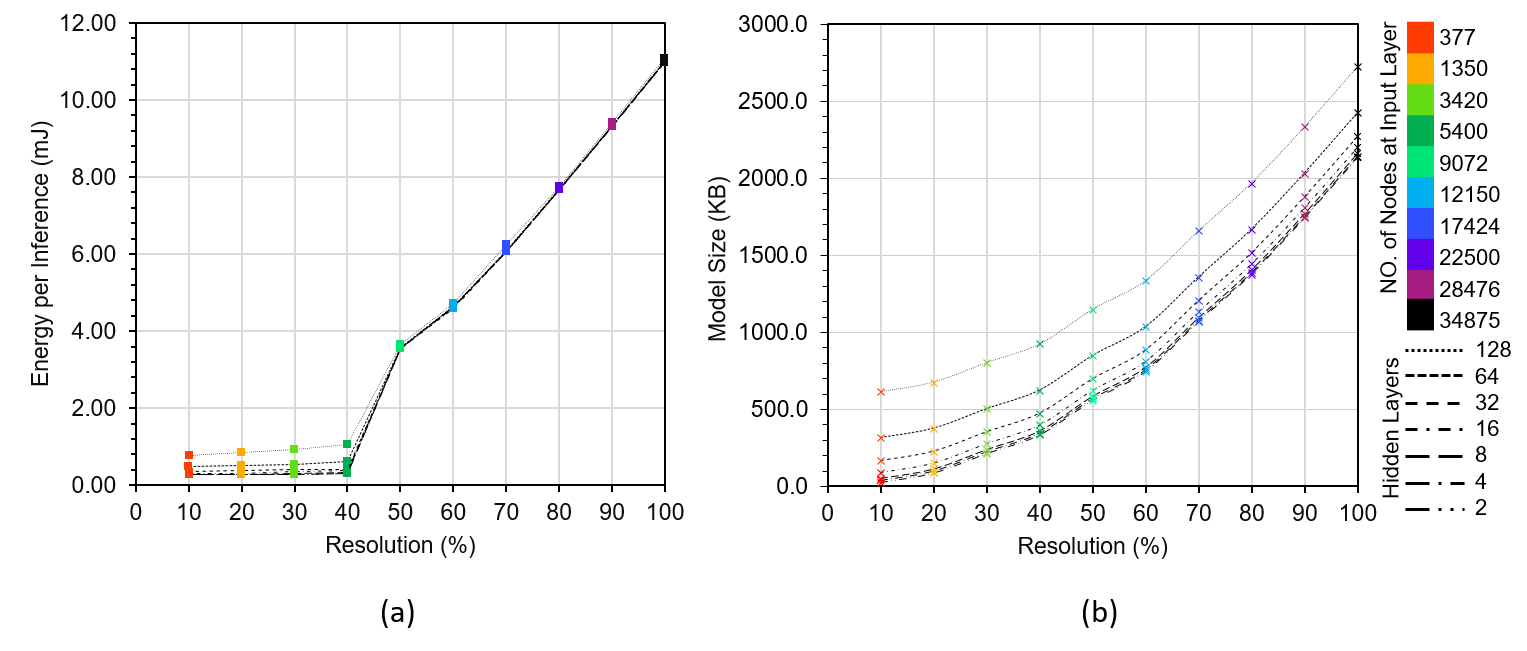}
\caption{\bm{Energy per inference on left (a) and model size on right (b) vs different spectrogram resolutions for Edge TPU. Different line markers show different number of hidden layers and the colour bar shows different number of input parameters which maps to the spectrogram resolutions.}}
\label{fig:Energy_TPU_res}
\end{figure*}

\orm{From a different point of view, Figure~\ref{fig:Energy_TPU_res}(a) shows the energy consumption of Edge TPU versus number of input nodes (spectrogram resolution). Here, we observe the bimodal behaviour of Edge TPU for regardless of number of  hidden layers,} with an almost constant and very low energy usage of 0.29~mJ per inference for the resolutions of 10-40\% and \orm{two hidden layers,} and an almost linear increase for the higher spectrogram resolutions, i.e. 50-100\%.
\orm{The same behaviour is observed for other steps of hidden layers with a slightly different range of 0.78-1.05~mJ per inference for the resolutions of 10-40\% and 128 hidden layers, and the transition happens for the resolutions beyond 40\%.}
\orm{In Figure~\ref{fig:Energy_TPU_res}(b) the size of all models remains below 8~MB.}


\begin{figure}[t!]
\centering
\includegraphics[width=9 cm]{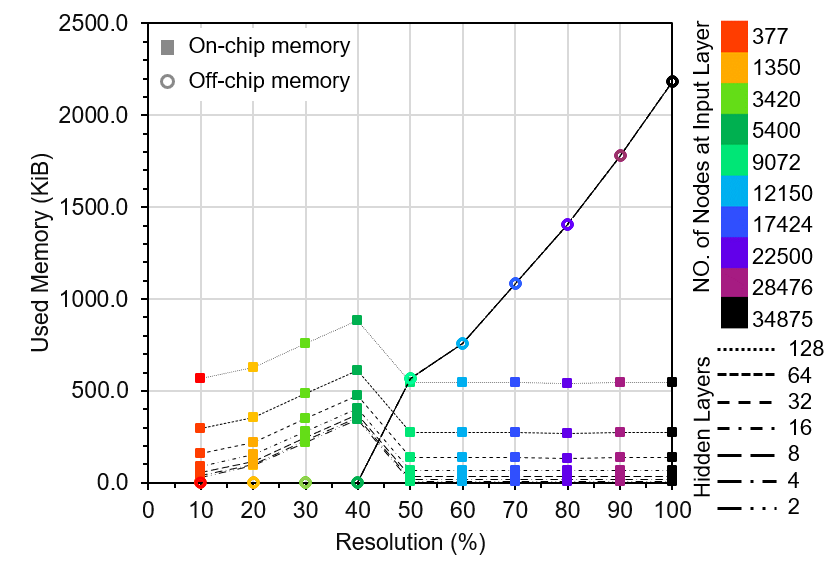}
\caption{\bm{Used off-chip and on-chip memory vs different spectrogram resolutions for Edge TPU. Different line markers show different number of hidden layers and the colour bar shows different number of input parameters which maps to the spectrogram resolutions.}}
\label{fig:Memory_TPU}
\end{figure}

%

\orm{The results of investigations in the compilation reports are illustrated in Figure~\ref{fig:Memory_TPU}.} This figure shows Edge TPU memory usage in KB, shown separately for on-chip and off-chip memory in KB, for the range of \bm{number of input nodes (spectrogram resolutions) and number of hidden layers.
We observe that regardless of number of hidden layers for number of input nodes 377 up to 5400 (a resolution of 10\% up to 40\%)} the use of on-chip memory linearly increases, while the off-chip memory is unused.
In this range, the entire neural network model and all its parameters fit into the on-chip memory, resulting in a highly efficient operation. This is clearly consistent with the shown energy per inference curve (see Figure~\ref{fig:Energy_TPU_res}(a)), which shows a very minimal and almost constant energy consumption in that range \bm{from 0.29~mJ to 1.05~mJ per inference.}
%
%
\orm{When the number of input nodes 
increases from 5400 to 9072, 
we observe that the on-chip memory usage drops by 76\% on average for different number of hidden layers. 
The rest of the required memory is provided with a corresponding use of off-chip memory. The amount of off-chip memory is fixed for each of different number of input nodes 
regardless of number of hidden layers or the model size.} 

\begin{figure}[t!]
\centering
\includegraphics[width=7 cm]{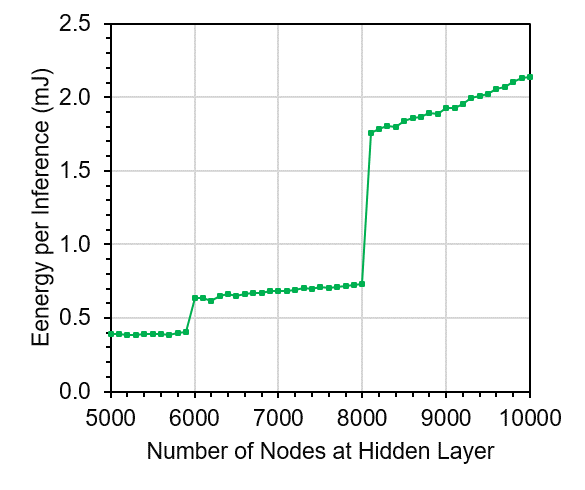}
\caption{Energy per inference vs number of nodes at hidden layer 
for configurations in Table\ref{tab:set3} on Edge TPU.
}
\label{fig:EnergyvsNodes_R10}
\end{figure}

\bm{Increase of the off-chip memory happens in steps, which the pattern matches the quantised pattern of energy usage observed in Figure~\ref{fig:EnergyvsNodes_R10}.
Here, the model size is far less than 8~MB (See Figure~\ref{fig:Energy_TPU_res}(b)).}
It seems the storage of model parameters in the on-chip memory follows an ``all-or-nothing" approach, with storage of the entire model in on-chip memory if it fits, and storage of the entire model off-chip, if not.

\begin{table}[b!]
  \centering
  \caption{\bm{Summary of the considered hyper-parameters to investigate the sensitivity of Edge TPU to number of parameters per layer}}
  \small
    \begin{tabular}{|l|l|}
    \hline
    Parameter & Range \\
    \hline
    $L$     &  2 \\
    \hline
    $n_{x}$ & {377} \\
    \hline
    \begin{tabular}[l]{@{}l@{}} $n_{h}^{[1]}$  \end{tabular} &  \{5000, 5100, 5200, ..., 10000\} \\
    \hline
     \begin{tabular}[l]{@{}l@{}} $n_{h}^{[2]}$  \end{tabular} &  32 \\
    \hline
    $n_{y}$ & 9 \\
    \hline
    \end{tabular}%
  \label{tab:set3}%
\end{table}%

\orm{So far, the role of model size and number of input parameters in the use of energy consumption is demonstrated.}
\grnm{To study the effect of number of parameters at hidden layers on the performance of the Edge TPU, we investigated the smallest on-chip models according to results shown in Figure~\ref{fig:EnergyvsModelsize_group} with $n_{x} = 377$ and $L = 2$. The exploration focuses on the number of nodes at the first hidden layer ($n_{h}^{[1]}$) ranging from 5000 to 10,000 with the step of 100. The number of nodes at the last hidden layer ($n_{h}^{[2]}$) and the output layer ($n_{y})$ are fixed at 32 and 9, respectively. The summary of the range hyper-parameters is presented in Table~\ref{tab:set3}.}
%

\begin{figure}[b!]
\centering
\includegraphics[width=7.5 cm]{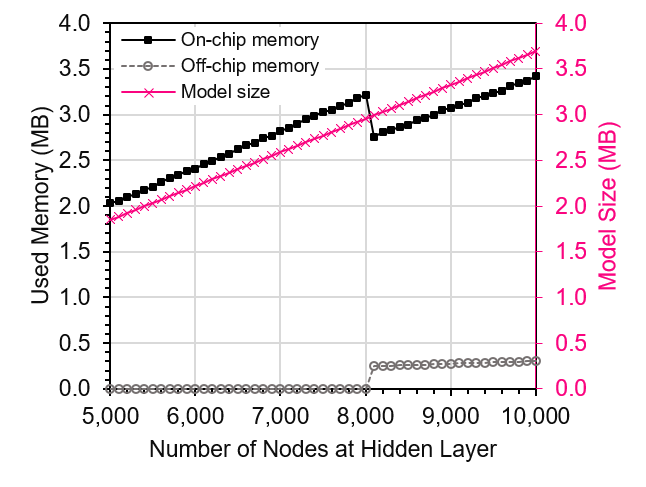}
\caption{\bm{On-chip and Off-chip memory usage and model size vs number of nodes at hidden layer for configurations in Table\ref{tab:set3} on Edge TPU. 
}}
\label{fig:MemoryvsNodesvsSize_R10}
\end{figure}

\redm{Figure~\ref{fig:EnergyvsNodes_R10} shows energy per inference vs number of nodes at the first hidden layer. Almost constant and very low energy usage of 0.39~mJ per inference for 5000 to 5900 nodes is observed, and 
at 6000 nodes a small step is noticed which increases the energy by 38\% to 0.63~mJ. The reason of this jump is not clear but it is assumed it is related to the hardware architecture of Edge TPU.}

\redm{Between 6000 and 8000 nodes, the energy usage increases linearly to 0.72~mJ. \orm{Then at 8000 nodes a significant step is observed which increases the energy usage by 58\% to 1.72~mJ. Beyond 8100, the energy usage increases linearly and reaches to 2.13~mJ.} 
The compilation reports for the models described above are presented in Figure~\ref{fig:MemoryvsNodesvsSize_R10}, shown the on-chip, off-chip memory usage, and model size. It is shown that increasing the nodes linearly increases the model size from 1.85~MB to 3.69~MB. It is interesting that at 8100 nodes, the transition happens and Edge TPU starts to use off-chip memory; however the model size is relatively smaller than the available on-chip memory.
\orm{We believe the reason is the amount of parameter memory of Edge TPU which is equal to 8192~\footnote{The amount of on-chip parameter memory is reported in~\cite{edgetpu2021}.}.
%
This explains the significant gap observed in Figure~\ref{fig:EnergyvsHiddenLayers_tpu} between 5400 and 9072 input nodes,
and also the significant jump in Figure~\ref{fig:EnergyvsNodes_R10} at 8000 nodes
i.e. the bimodal behaviour of the Edge TPU.
Hence, number of nodes per layer is determinant factor for energy usage as well as the model size. The number of nodes per layer should stay fewer than 8000 for the model to fit into on-chip available memory.}}

\prpm{The Edge TPU compiler determines the mapping of data based on the available on-chip memory (processing engine memory and core memory), the model parameters (input activations, weight parameters), and outputs.}
\prpm{The Edge TPU ASIC is build up on 2D arrays of processing engines (PE) which work on a Single Instruction Multiple Data (SIMD) logic. 
Inside a PE, multiple or single compute lanes and the shared PE memory are engaged in processing a set of activations that are given in each cycle. 
The scarce on-chip memory including the PEs memory and the core memory are used for storing input activations, partial sums, outputs of computations, and weight parameters.
Based on SIMD, caching model parameters provides efficiency by reducing the parameter transfers to be used in each cycle~\cite{edgetpu2021}. 
However, when the model size or number of parameters exceeds the on-chip memory limit the transition happens, and the off-chip memory is used for streaming model parameters resulting to the bimodal behaviour of Edge TPU. Thus, the IO bandwidth becomes the critical factor.
}
\redm{The bimodal behaviour of the Edge TPU can be triggered by the number of parameters per layer when greater than 8000 or when the model size is greater than 8~MB. Hence, to stay on-chip, the model specifications should stay below these limitations.}


\subsection{Energy Efficiency Ratio of Edge TPU over Cortex-A53}\label{res-ratio}

\begin{figure}[b!]
\centering
\includegraphics[width=7.5cm]{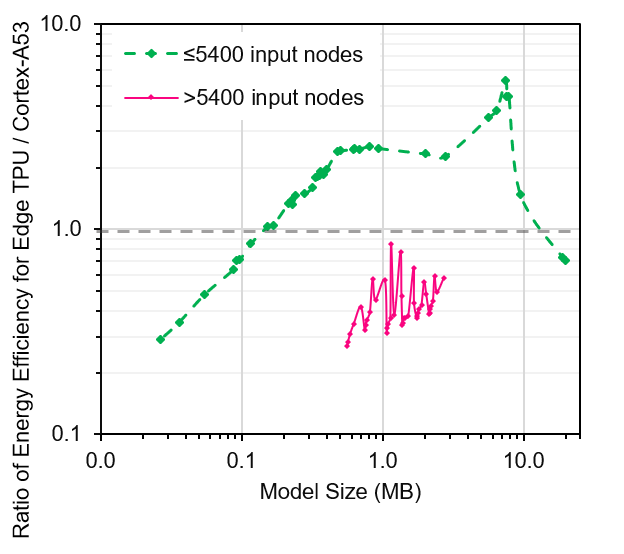}
\caption{\bm{The ratio for energy efficiency of Edge TPU over Cortex-A53 vs model size; focusing on number of input nodes.}}
\label{fig:Ratio}
\end{figure}

In machine learning (ML) at the edge, both performance, e.g. classification $F_1$ Score in our use case, and energy efficiency are critical.
So far, we have considered classification performance 
\orm{and energy efficiency vs different variations of DNN architectural design space including number of input nodes, number of hidden layers and number of nodes at hidden layers. We have seen there is no significant difference in classification performance of HW platforms. However, the energy efficiency of the Edge TPU is bimodal and is dependent on both the model size and number of parameters per layer. This bimodal behaviour caused Cortex-A53 with its linear behaviour, to outperform Edge TPU at some points.}  

\orm{In order to simplify the comparison of the Cortex-A53 and Edge TPU,  
we summarise \bm{the results by considering 
the ratio of energy efficiency of Edge TPU over Cortex-A53.} 
Figure~\ref{fig:Ratio} shows the ratio on y-axis and model size (MB) on x-axis.
It is observed that for models with fewer than 5400 input nodes when the model size is less than 0.15~MB, Cortex-A53 is more efficient than the Edge TPU. Then, with the increase in the model size, the Edge TPU outperforms Cortex-A53. The advancement of the Edge TPU is at highest before the model size exceeds 8~MB; and Cortex-A53 overtakes Edge TPU at 13.5~MB. 
However, for greater than 5400 input nodes regardless of the model size Cortex-A53 outperforms Edge TPU.
It is important to note that, Edge TPU is highly sensitive to the choice of model size and number of parameters per layer.}


\section{Design Considerations to use Edge TPU}\label{guideline}

Our investigations show that great care needs to be taken when designing neural network models for Edge TPU. If done right, and with a model size that fits on the on-chip memory, an excellent energy performance and efficiency can be achieved. Conversely, if the model is too large and does not fit in the on-chip memory, the energy efficiency of the Edge TPU is not much better than that of traditional CPUs. For practical ML applications at the edge, it is therefore critical to have the ability to scale the model size to the ``sweet spot", which provides both good performance and energy efficiency. 


\begin{figure}[b!]
\centering
\includegraphics[width=8.5cm]{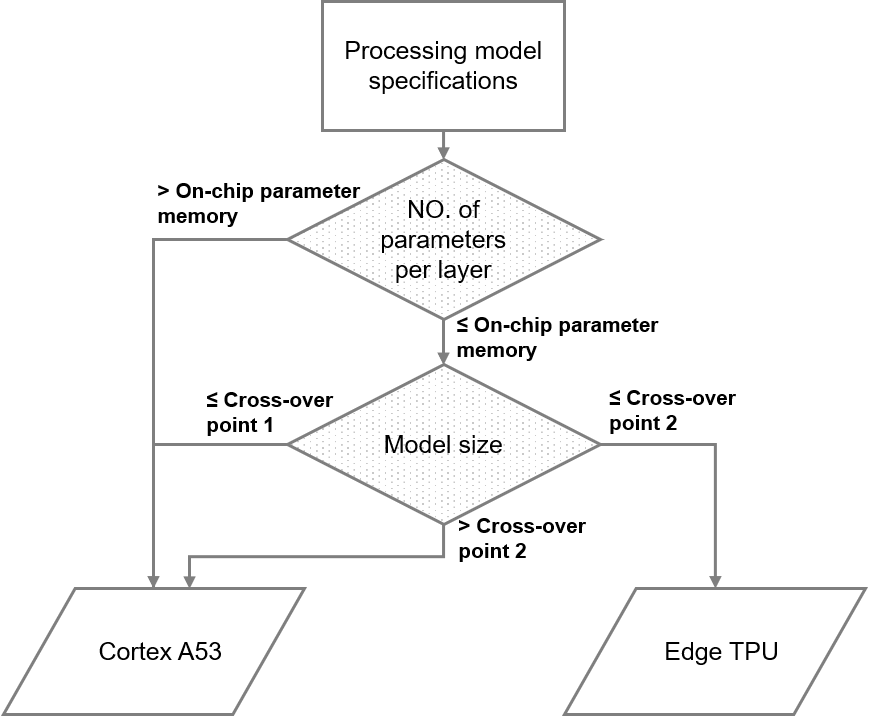}
\caption{\bm{Suggested co-design for feed forward neural networks implementation on Edge TPU.}}
\label{fig:guideline}
\end{figure}

\bm{The Edge TPU specifications, at the time of this study, are 8~MB on-chip memory and 8192 on-chip parameter memory. And according to the findings of this study, Edge TPU outperforms Cortex-A53 between two points, which we refer to as the cross-over point 1 at 0.15~MB and the cross-over point 2 at 12.5~MB. In this regard, the key points suggested to consider when designing a feed forward neural network to be implemented on Edge TPU are listed below:
\begin{itemize}
  \item{If the model is small with the model size less than the cross-over point 1, Cortex-A53 is more efficient than Edge TPU.} 
  \item{If the model size is less than the available on-chip memory (8~MB) and number of parameters per layer is fewer than the on-chip parameter memory (8192), Edge TPU performs at its sweet spot.}
  \item{If number of parameters per layer is more than the on-chip parameter memory, Cortex-A53 is more efficient.}
  \item{If the model size is greater than the cross-over point 2, then cortex-A53 is more efficient.} 
\end{itemize}
}
\orm{Figure~\ref{fig:guideline} summarises the above points in a flowchart for feed forward neural networks implementation on Edge TPU}.

\section{Conclusions}\label{conclusion}

In this paper, we have \bm{evaluated the} Google's Edge TPU for \bm{deep feed forward neural network }
at the edge, and compared its performance with that of \bm{both traditional Intel and embedded 64-bit ARM CPUs.} 
For our exploration, we considered the use case application of activity classification in particular cattle, which requires inference to run at the edge, on highly resource-constrained embedded systems. 
\orm{A key focus of this paper was the exploration of the DNN model architectural variations.
In our experiments we used the spectrogram data representation to scale the input size of DNN, as} it has excellent scaling properties, i.e. it allows the compression of the data to a fraction of its original size, without sacrificing much of its predictive power and hence classification accuracy in the context of our considered application. 
\bm{Further, the hyper parameters of the DNN architecture which leads to scaling the model size and \orm{number of parameters per layer} are explored for the feed forward neural network.}

Our experimental results have shown the Edge TPU can provide excellent classification accuracy and energy efficiency that is significantly higher than traditional CPUs, and hence has great potential for application in energy constrained IoT devices.
However, we have also shown that the Edge TPU performance and energy efficiency is highly sensitive to the neural network model size \bm{and number of parameters. If the model size or number of parameters per layer} is too large to fit on the on-chip TPU memory, the energy efficiency significantly drops to a level that is almost on par with the considered CPUs. \bm{Surprisingly, for very small neural network models Cortex-A53 is faster and more energy efficient than Edge TPU.}


While our exploration was based on a specific use case application and a specific edge ML hardware platform, we believe the results are applicable more generally in the context of ML at the edge. 

\section*{Acknowledgments}
We acknowledge the support of the following researchers in regards to data collection: Greg J. Bishop-Hurley with CSIRO Agriculture and Food, and Paul Greenwood, Alistair Donaldson and Reg Woodgate  with NSW Department of Primary Industries, and Jody McNally , Troy Kalinowski  and Aaron Ingham with CSIRO Agriculture and Food. We also acknowledge support from Steffen Bollmann with centre of advanced imaging at the university of Queensland.

\section*{}
\bibliographystyle{IEEEtran} 
\bibliography{References} 

\end{document}